%% file: main.tex
\documentclass[conference]{IEEEtran}
\usepackage{times}

\usepackage[numbers]{natbib}
\usepackage{multicol}
\usepackage[bookmarks=true]{hyperref}
\usepackage{amsmath}
\usepackage{amssymb}
\usepackage{textcomp}
\usepackage{url}
\usepackage{verbatim}
\usepackage{graphicx}
\usepackage{booktabs}
\usepackage{multirow} 
\usepackage{xcolor}
\usepackage[caption=false]{subfig}

\begin{document}

\title{Hierarchical Audio-Visual-Proprioceptive Fusion for Precise Robotic Manipulation}




%
\author{
\IEEEauthorblockN{
Siyuan Li$^{1*\dagger}$,
Jiani Lu$^{1*}$,
Yu Song$^{1}$, 
Xianren Li$^{1}$, 
Bo An$^{2}$, 
Peng Liu$^{1}$
}

\IEEEauthorblockA{$^{1}$Harbin Institute of Technology, China}
\IEEEauthorblockA{$^{2}$Nanyang Technological University, Singapore}

\IEEEauthorblockA{$^{*}$Equal contribution}
\IEEEauthorblockA{$^{\dagger}$Corresponding author: siyuanli@hit.edu.cn}
}

\maketitle

\begin{abstract}
Existing robotic manipulation methods primarily rely on visual and proprioceptive observations, which may struggle to infer contact-related interaction states in partially observable real-world environments.
Acoustic cues, by contrast, naturally encode rich interaction dynamics during contact, yet remain underexploited in current multimodal fusion literature.
Most multimodal fusion approaches implicitly assume homogeneous roles across modalities, and thus design flat and symmetric fusion structures. However, this assumption is ill-suited for acoustic signals, which are inherently sparse and contact-driven.
To achieve precise robotic manipulation through acoustic-informed perception, we propose a hierarchical representation fusion framework that progressively integrates audio, vision, and proprioception. 
Our approach first conditions visual and proprioceptive representations on acoustic cues, and then explicitly models higher-order cross-modal interactions to capture complementary dependencies among modalities. 
The fused representation is leveraged by a diffusion-based policy to directly generate continuous robot actions from multimodal observations. 
The combination of end-to-end learning and hierarchical fusion structure enables the policy to exploit task-relevant acoustic information while mitigating interference from less informative modalities. 
The proposed method has been evaluated on real-world robotic manipulation tasks, including liquid pouring and cabinet opening. Extensive experiment results demonstrate that our approach consistently outperforms state-of-the-art multimodal fusion frameworks, particularly in scenarios where acoustic cues provide task-relevant information not readily available from visual observations alone. 
Furthermore, a mutual information analysis is conducted to interpret the effect of audio cues in robotic manipulation via multimodal fusion.

\end{abstract}

\IEEEpeerreviewmaketitle

\section{Introduction}
\begin{figure}[!t]
\centering
\includegraphics[width=\linewidth]{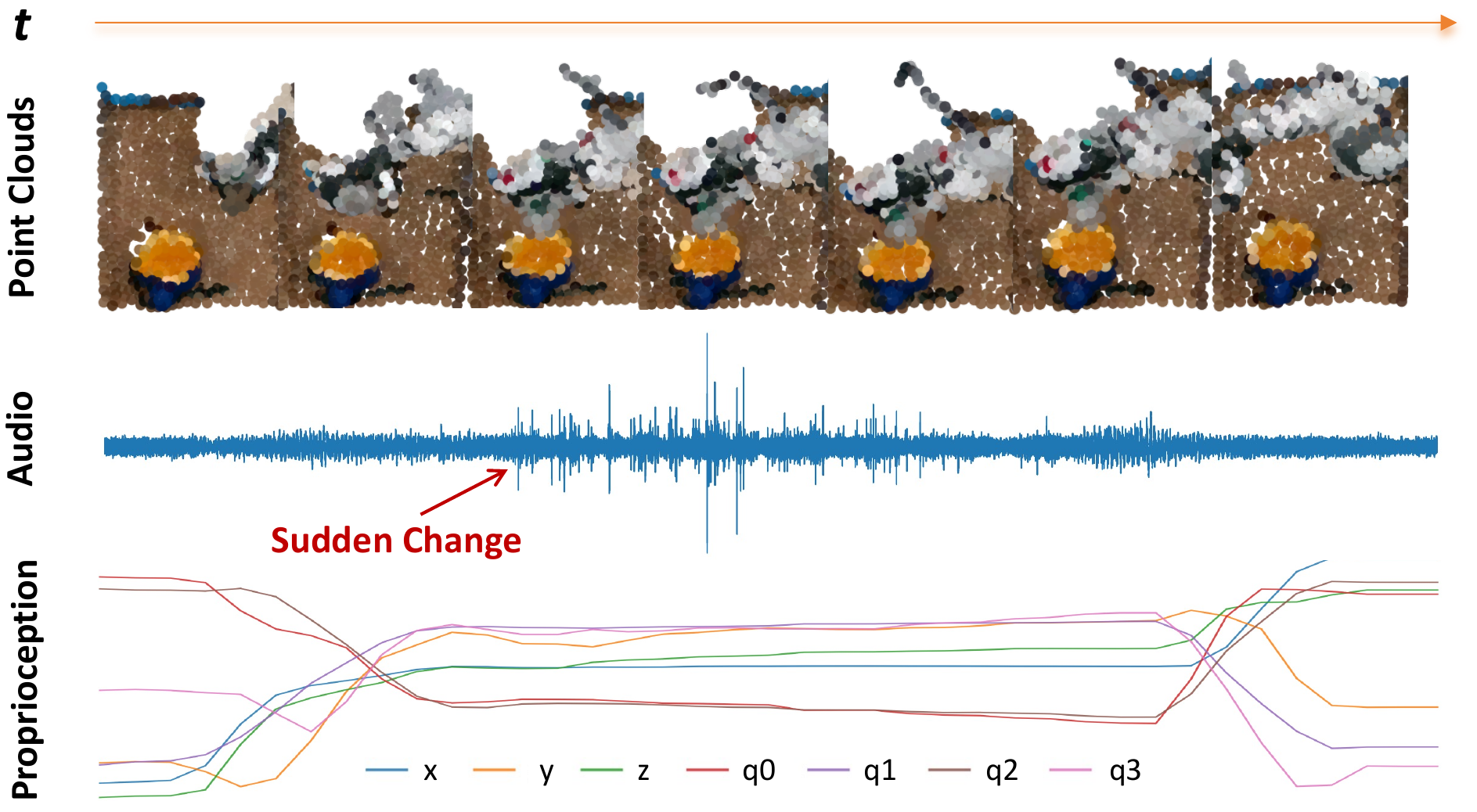}
\caption{{\bf Example trajectory of visual, audio, and proprioception observations in a real-world pouring task. }
The three modalities are temporally synchronized, yet exhibit markedly different characteristics.
Visual and proprioceptive signals vary smoothly over time, reflecting gradual motion and pose changes, whereas audio signals are sparse and exhibit abrupt transients tightly coupled with physical interactions.
These empirical observations suggest that acoustic information encodes interaction-specific cues that are complementary to vision, and these three modalities are inherently hetergeneous.}
\label{fig:firstFig}
\end{figure}
Robotic manipulation is a prominent research direction in embodied intelligence, and most existing manipulation work relies on two types of perception \cite{An-24-RGBManip,faroni2025uncertaintyawareplanninginaccuratemodels,Lin_2023_ICCV,Chi-RSS-23,Manawadu-mdpi-24, Ze2024DP3}: visual external sensing to perceive the environment and proprioceptive sensing to perceive robots' own state. 
Although visual observations provide detailed spatial information about surrounding environments, 
purely visual observations are insufficient to fully capture the dynamics and rich contact during physical interaction, hindering precise and robust manipulation. 
In particular, visual observations are sensitive to occlusion and struggle with transparent or reflective objects~\cite{10175024,10288041}, which commonly appear in real-world environments. 
In contrast, acoustic sensing is promising for capturing fine-grained interaction dynamics that are challenging or impossible to perceive visually~\cite{10889950, liang2019AudioPouring,liang2020MultimodalPouring}. 
As audio signals are able to encode subtle temporal dynamics induced by physical contacts and material properties, audio is regarded as complementary information that effectively reflects underlying interaction states beyond the visual modality~\cite{GUO2023104271,thankaraj2023that}. 
Therefore, integrating visual, acoustic, and proprioceptive modalities is promising for achieving precise robotic manipulation in real-world environments.

Existing audio-visual manipulation approaches either handle audio signals with additional supervision labels \cite{Wang-ITNNLS-25} and task-specific knowledge \cite{Ruoxuan-play-2024}, or perform multimodal fusion with lightweight structures, such as concatenation \cite{Ye-MORPHeus-2024} or basic attention \cite{li2022seehearfeel}. More recent works try to design sophisticated architectures for more effective fusion \cite{Ye-MORPHeus-2024,Zhang-ITASE-24,MejiaiICRA24}, and ManiWAV \cite{pmlr-v270-liu25c} stands out as a superior method among them, as it can leverage in-the-wild audio signals.
In most previous work, audio is treated as a homogeneous modality with visual and proprioceptive observations.
However, these modalities are inherently heterogenous.
As illustrated in Fig. \ref{fig:firstFig},
audio signals are contact-driven, sparse, and vary abruptly. In contrast, the variation of visual observation is much smoother.
The substantial differences among audio, visual, and proprioceptive observations constrain the effectiveness of symmetrically designed fusion structures that treat all modalities equally.

To address this heterogeneous fusion issue, we propose
a novel hierarchical audio-visual-proprioceptive fusion approach that especially emphasizes the audio modality.
Different from symmetrical fusion methods, we first utilize the audio modality to augment the other two separately through a \textit{binary-branched fusion module}. Then, we obtain two fused representations: an audio-visual representation and an audio-proprioceptive representation. 
These two representations extract cross-modal features, but due to the sparsity of audio signals, contact-related information valuable for precise manipulation may be attenuated. 
To emphasize audio more, we build a highway from the audio encoder to a fusion module that further models cross-modal interactions, called \textit{interaction modeling module}.
These two modules (binary-branched fusion and interaction modeling) compose the hierarchical multimodal fusion framework, 
 and the resulting fused representation is finally used within an end-to-end diffusion-based manipulation policy to generate continuous actions.

 The proposed approach is evaluated in real-world manipulation tasks, including liquid pouring and cabinet opening, where precision is critical.
The experimental results show that our approach consistently outperforms baselines that rely on single modalities or state-of-the-art audio-visual fusion strategies, demonstrating better precision and stronger generalization. 
Furthermore, we conduct a mutual information analysis and ablation studies to probe the proposed audio-centric hierarchical fusion, revealing that audio information plays a dominant role in acoustically rich tasks.
The main contributions of this paper can be summarized as follows:
\begin{itemize}
\item A binary-branched fusion module that uses audio as a conditioning signal to modulate visual and robot proprioception representations.
\item An interaction modeling module built upon an audio highway, which further models the higher-order cross-modal interactions with an emphasize of audio signals. 
    
    
    \item Extensive experiment results not only demonstrate the precision and generalization of multimodal manipulation, but also validate the effectiveness of each module.
\end{itemize}

\section{Related Work}
This section briefly reviews the robotic manipulation literature about the two external sensing modalities, vision and audio, and their fusion. Specifically, Section~\ref{rw:a} reviews visual and audio representation learning in isolation, and Section~\ref{rw:b} discusses the audio-visual fusion works.

\subsection{Visual and Audio Representations}
\label{rw:a}
Vision-based observations, such as RGB~\cite{An-24-RGBManip, faroni2025uncertaintyawareplanninginaccuratemodels,Lin_2023_ICCV} or RGB-D images~\cite{Chi-RSS-23,Manawadu-mdpi-24}, provide rich perceptual information for robotic manipulation, such as geometric, spatial, and semantic properties of surrounding objects~\cite{10.5555/3618408.3619019}.
In recent years, point clouds have emerged as a powerful visual modality, which has demonstrated greater spatial generality than RGB images.
Early visual-based approaches typically adopt point-based encoders, such as PointNet~\cite{CharlesPointNet2017} and its variants~\cite{QiPointNetplusplus2017,qian2022pointnext}. 
These encoders have been widely used in manipulation policies and combined with robot proprioception or language instructions to predict actions, as demonstrated in works such as DexPoint \cite{qin2022dexpoint}, PolarNet \cite{chen2023polarnet}, and PointFlowMatch \cite{chisari2024learning}. 
However, aggressive pooling operations often compress rich spatial information
into a global representation, limiting the expressiveness for fine-grained
interaction reasoning.
To improve efficiency and training stability, more recent works adopt lightweight point cloud encoders that focus on extracting compact, task-relevant geometric features~\cite{Ze2024DP3,yan2025maniflow}.
In parallel, other works explore structured point cloud representation, such as voxel grids~\cite{Geng_2023_CVPR}, sparse convolutional features, or point maps that rearrange 3D points into regular girds~\cite{jia2025pointmappolicy}, enabling the use of convolutional or transformer-based architectures at the
cost of increased preprocessing complexity. 
Therefore, in this work, we adopt point clouds as the visual perception. 

Inspired by humans' ability to infer states from acoustic cues in visually partially observable environments, previous research attempts to leverage audio signals in robotic pouring to estimate liquid height~\cite{10889950,liang2019AudioPouring,liang2020MultimodalPouring}. 
Beyond state estimation, \citet{GUO2023104271} propose to utilize interaction sounds for object recognition and instruction understanding.
\citet{thankaraj2023that} develop a self-supervised framework that uses audio to generate robot actions, but relying solely on audio lacks spatial and geometric context.
Generally, most audio representation learning works in robotic manipulation serves as one component in the perception-decision-action pipeline, but has not been integrated into the end-to-end policy learning.
These methods contribute to audio feature extraction for the following audio-visual fusion methods.

\subsection{Audio-Visual Fusion}
\label{rw:b}
Audio–visual fusion has been extensively explored across various robotic perception tasks, demonstrating the strong complementarity between acoustic and visual modalities. 
Early related research adopts lightweight fusion structures, e.g., simple concatenation of audio and visual representations~\cite{Ye-MORPHeus-2024} or basic attention mechanism \cite{li2022seehearfeel}.
Without a well-designed fusion strategy, these methods can hardly
leverage the unique strengths of each modality effectively.
To better support robotic manipulation, more recent audio-visual fusion works try to develop more sophisticated architure.
For example, considering the long-term temporal history, \citet{Maximilian-RSS-22} propose to concatenate audio, visual, and robot state embeddings before processing them through a Multi-Layer Perceptron (MLP) to obtain a joint feature.
The following works develops staged or task-specific fusion pipelines \cite{Zhang-ITASE-24,Ruoxuan-play-2024}, or rely on additional supervision to learn intermediate semantic representations~\cite{Wang-ITNNLS-25}. 
One of the most related works is ManiWAV~\cite{pmlr-v270-liu25c}, which concatenates audio and visual features and fuses them using a Transformer encoder. Concurrently to ManiWAV, ~\citet{MejiaiICRA24} also propose a similar Transformer-based fusion framework. 
In spite of the progress in audio-visual fusion, most related work assumes that the visual and audio modalities are homogeneous, neglecting the contact-driven and sparsity property of audio signals. 
Therefore, the critical information for manipulation from audio signals has been underexploited.
To address this issue, we propose a novel hierarchical audio-visual-proprioceptive fusion framework that especially emphasizes the audio modality.

\section{Problem formulation and preliminaries}
\subsection{Problem Formulation}
We consider a robotic manipulation task in which observations are obtained from three types of sensing: visual, audio, and proprioceptive.
The learning objective of this work is to derive a policy that maps raw observations to robot actions. 
The robotic manipulation is temporally extended, and at each timestep $t$, the multimodal observation $o_t$ is denoted as follows:
\begin{equation}
    o_t=\{o_t^a,o_t^p,o_t^s\},
\end{equation}
where $o_t^a\in \mathbb{R}^{N_o \times T\times M}$ is the audio segment that starts at timestep $t-T$, and ends at timestep $t$. $T$ serves as a sliding window size to capture the temporally related information, and $M$ corresponds to the number of Mel-frequency bins for the audio signals.
$o_t^p\in \mathbb{R}^{N_o \times N \times 3}$ represents the 3D spatial point cloud observation, and $N$ is the point number of each frame. $o_t^s\in \mathbb{R}^{N_o \times D_s}$, where $D_s$ indicates the dimension of the robot proprioception. 
$N_o$ denotes the number of recent timesteps stacked along the temporal dimension, providing necessary temporal context, as a single observation is insufficient to capture robot motion dynamics.
Since the raw observations $\{o_t^a,o_t^p,o_t^s\}$ are generally high-dimensional, these observations are first processed by the corresponding feature encoders before the fusion process. 
\begin{equation}
    z_t=F(f_a(o_t^a),f_p(o_t^p),f_s(o_t^s)),
    \label{eq_z}
\end{equation}
where $f_a,f_p,f_s$ represent the encoders of audio, point cloud, and robot proprioception separately, and the ouputs of these encoders are denoted as $\{x_t^a, x_t^p, x_t^s\}$.
The main focus of this work is to develop the fusion function $F$ to obtain the fused latent representation $z_t$ that 
fully captures the acoustic dynamic, the spatial scene geometry, and the robot motion.

In this work, we adopt an imitation learning method \cite{hussein2017imitation} to optimize the manipulation policy, so we assume an expert dataset $\mathcal{D}_{exp}$ is given, which contains $N_{exp}$ successful trajectories for the target task.
The policy $\pi$ takes the latent embedding $z_t$ in Equation \eqref{eq_z} as input, and outputs the robot action. The robot action $a$ is defined over the end-effector position-orientation space, which is standard in the robotic manipulation literature. 
The whole manipulation policy containing the feature encoders $\{f_a,f_p,f_s\}$, the fusion network $F$, and the control policy network $\pi$ is optimized in an end-to-end manner with behavior cloning, and the details about policy $\pi$ are presented in the following subsection. 

\subsection{Diffusion Policy}
\label{pre}
Diffusion models \cite{yang2023diffusion} are widely used in sequence generation, which have also been adopted for action generation in robotic manipulation.
Following the DP3 approach \cite{Ze2024DP3}, we employ a diffusion-based policy to generate future robot actions by modeling a conditional distribution over action trajectories.
Let $\mathbf{a}_{t:t+H} = \{a_t, \dots, a_{t+H}\}$ denote the future actions to execute with a horizon of $H+1$, where $a_t \in \mathbb{R}^{D_a}$ is the end-effector control command. 
The output action is sampled from the distribution
\begin{equation}
  \mathbf{a}_{t:t+H} \sim  \pi( \cdot \mid z_t ;k),
\end{equation}
where $z_t$ is the multimodal latent feature in Equation \ref{eq_z}, and $k$ indicates the noise level in the diffusion network.

The entire policy network is optimized with the following behavior cloning loss function:
\begin{equation}
    \mathcal{L}_{\text{diff}} =
    \mathbb{E}_{\mathcal{D}_{exp}}
    \left[
    \left\|
    \mathbf{a}_{t:t+H} -
    \hat{\mathbf{a}}_{t:t+H}
    \right\|_2^2
    \right],
    \label{loss}
\end{equation}
where $\hat{\mathbf{a}}$ denotes the action sequence sampled from the expert dataset $\mathcal{D}_{exp}$, and $\mathbf{a}_{t:t+H}$ is the output of the diffusion policy with the corresponding observations. Note that all the parameters in the policy network, including diffusion-based $\pi$, the feature encoders $\{f_a,f_p,f_s\}$, and the fusion network $F$, are all optimized with the loss function in Equation \eqref{loss}. 
At inference time, an action sequence is generated by iterative denoising conditioned on $z_t$.
Only the last $N_a$ actions will be executed, and the process is repeated in a receding-horizon manner.

\section{Method}  
\begin{figure*}[!t]
\centering
\includegraphics[width=0.9\textwidth]{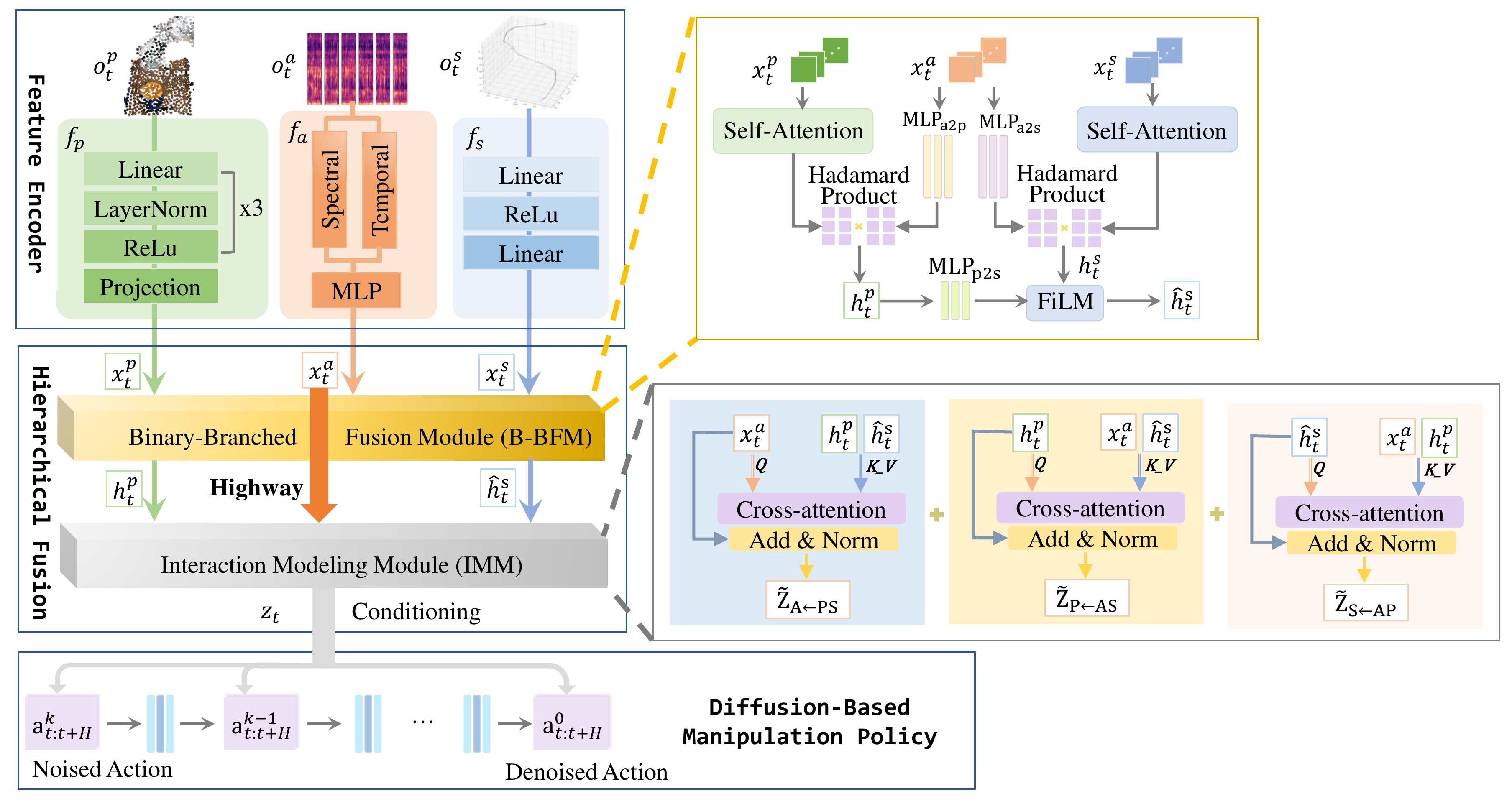}
\caption{{\bf The architecture of the proposed hierarchical fusion method.} Each modality is initially encoded, followed by the Binary-Branched Fusion Module, which aggregates the extracted features. The resulting intermediate features then undergo interactive fusion to yield the final embedding $z_t$, which serves as the conditioning input to the diffusion policy. This policy iteratively denoises the trajectory from random noise to executable robot actions.}
\label{fig:archiFig}
\end{figure*}

To achieve effective multimodal fusion with an emphasize of the sparse and fluctuating audio signals, we propose a novel hierarchical audio-visual-proprioceptive fusion architecture, as illustrated in Fig.~\ref{fig:archiFig}. The proposed hierarchical architecture consists of two main components: a binary-branched fusion module (B-BFM), which aggregates these modalities progressively, described in Section \ref{subsec:B}, and an interaction modeling module (IMM) presented in Section \ref{subsec:C}, using a cross-attention mechanism to model the higher-order interactions among these modalities with an emphasize of audio signals. As a precondition for multimodal fusion, we employ separate feature encoders for each modality to process the high-dimensional raw observations, as depicted in Section \ref{subsec:A}.

\subsection{Separate Feature Encoder}  
This subsection elaborates on the feature encoders $\{f_a,f_p,f_s\}$ for the audio, point cloud, and proprioception observations. As illustrated in the upper part of Fig. \ref{fig:archiFig},
The outputs of these encoders $\{x_t^a, x_t^p, x_t^s\}$ serve as preconditions of the following hierarchical fusion model.
\label{subsec:A}

\noindent \textbf{Audio Encoder.}
Audio signals contain rich information across both temporal and frequency dimensions.
Therefore, we propose to extract acoustic features using a dual-path audio encoder that captures complementary temporal and spectral cues from the log-Mel spectrogram.
Specifically, the encoder consists of two parallel branches.
The temporal branch $f_{\mathrm{time}}$ focuses on modeling temporal dynamics across time frames to capture fine-grained temporal variations induced by physical interactions, 
while the spectral branch $f_{\mathrm{freq}}$ emphasizes spectral patterns to extract frequency-dependent features associated with interaction-induced vibrations. 
The outputs of the two branches are concatenated to extract the audio embedding $x_t^a \in \mathbb{R}^{D}$ as follows:
\begin{equation}
\begin{aligned}
    x_t^a    &= f_a(o_t^a)\\   
    &=f_{\mathrm{enc}}^a\!\left(
    f_{\mathrm{time}}(\mathrm{LogMel}(o_t^a))
    \;\Vert\;
    f_{\mathrm{freq}}(\mathrm{LogMel}(o_t^a))
    \right),
    \end{aligned}
\end{equation}
where $\Vert$ denotes the concatenation of features, $D$ denotes the embedding dimension. $f_{\mathrm{enc}}^a$, $f_{\mathrm{time}}$ and $f_{\mathrm{freq}}$ constitute the audio encoder $f_a$.

Acoustic cues alone can be ambiguous for manipulation, as similar sounds may arise from distinct interaction events under different physical conditions.
To ground acoustic representations in the physical interaction context, we pretrain the audio encoder jointly with the proprioceptive encoder $f_s(o_t^s)$ using an autoencoding objective.
This joint pretraining encourages the audio encoder to learn representations that are physically grounded in the robot’s interaction state. 
The pretrained audio encoder is then used to extract acoustic features for subsequent audio–visual–proprioceptive fusion.

\noindent \textbf{Visual Encoder.} 
Inspired by the DP3 approach~\cite{Ze2024DP3}, we process the point clouds with an encoder composed of linear layers, layer normalization, ReLu activation, and a projection layer. Before being fed into the visual encoder $f_p$, the point cloud $o_t^p$
is cropped to the task-relevant region and downsampled to $N$ points using farthest point sampling, so that each visual observation maintains a unified size no matter how many points are captured by the camera. In addition, because the manipulation tasks considered in this work primarily depend on geometric structure rather than appearance, color information is discarded. 

\noindent \textbf{Proprioceptive Encoder.} As the dimension of the proprioceptive observation $o_t^s$ is lower than the other two modalities, its encoder is relatively lighter than the others. As shown in the upper right part of Fig. \ref{fig:archiFig}, the proprioceptive encoder $f_s$ is a two-layered MLP activated with the ReLu function, which output a proprioception embedding $x_t^s \in \mathbb{R}^{D}$.

\subsection{Binary-Branched Fusion Module}
\label{subsec:B}
In manipulation tasks, audio primarily conveys interaction events, while point cloud and proprioceptive inputs describe the environment geometry and the robot’s motion, respectively.
To extract the correlation between audio and the other two modalities,
we propose a binary-branched fusion module (B-BFM) that integrates audio, vision, and proprioceptive information through structured, feature-wise modulation. 
Rather than treating audio as an independent input, B-BFM uses it as a conditioning signal that modulates the interpretation of the other modalities.

As illustrated in the upper right of Fig.~\ref{fig:archiFig}, the module consists of two parallel binary branches. In the first branch, point cloud features are processed by a self-attention layer to capture spatial dependencies within the scene representation. In the last branch, proprioceptive features are similarly processed to model temporal and kinematic structure in the robot’s motion. In the middle, audio features are transformed by MLPs to produce branch-specific modulation signals, which are then applied via the Hadamard product. This operation injects event-related and phase-related information into visual and proprioceptive features, allowing geometric and motion features to be interpreted in the context of the current interaction:
\begin{subequations}
\begin{equation}
h_t^{p}= \mathrm{SelfAttn}(x_t^p) \odot \mathrm{MLP}_{a2p}(x_t^a),
\end{equation}
\begin{equation}
h_t^{s}= \mathrm{SelfAttn}(x_t^s) \odot \mathrm{MLP}_{a2s}(x_t^a),
\end{equation}
\end{subequations}
where $\odot$ denotes the Hadamard product.

To further integrate the two branches, we adopt a Feature-wise Linear Modulation (FiLM) mechanism \cite{perez2018film}. Specifically, the audio-modulated point cloud features are used to generate feature-wise affine parameters $\boldsymbol{\beta}$ and $\boldsymbol{\gamma}$ that modulate the proprioceptive represetation. Formally, given the intermediate proprioceptive representation $h_t^s$, the FiLM-modulated output is computed as 
\begin{equation}
\hat{h}_t^{s} = (1 + \boldsymbol{\gamma}) \odot h_t^{s} + \boldsymbol{\beta}.
\end{equation}
The affine parameters are generated from $h_t^{p}$ through a learnable mapping:
\begin{equation}
[\boldsymbol{\gamma}_t \;\Vert\; \boldsymbol{\beta}_t]
=
\mathrm{MLP_{p2s}}(h_t^{p}),
\end{equation}  
where $\mathrm{MLP}_{p2s}(\cdot)$ outputs a $2D$-dimensional vector that is split along the feature dimension to obtain the scale $\boldsymbol{\gamma}_t \in \mathbb{R}^{D}$ and shift $\boldsymbol{\beta}_t \in \mathbb{R}^{D}$.
This formulation enables geometric context, informed by acoustic events, to directly influence how robot motion states are represented.

Overall, the fusion process follows a binary-tree structure. Individual modality features form the leaves, pairwise fusion operations constitute intermediate nodes, and the final FiLM-modulated representation serves as the root. This hierarchical design allows the model to selectively exploit informative acoustic cues while suppressing irrelevant ones, leading to a unified multimodal embedding that captures both modality-specific structure and their interactions.

\subsection{Interaction Modeling Module}
\label{subsec:C}
While B-BFM integrates modalities in a structured and conditioned manner, it primarily captures pairwise dependencies. To further model higher-order interactions among modalities, we introduce an interaction modeling module (IMM) based on cross-attention as shown in the lower right of Fig.~\ref{fig:archiFig}.
Since audio signals are sparse and fluctuating, as illustrated in Fig. \ref{fig:firstFig}, to avoid missing the important information from audio signals, we additionally introduce a \textit{highway} that crosses B-BFM to directly pass the output of the audio encoder to the lower-level IMM.
Furthermore, in contrast to simple feature aggregation, IMM allows each modality to update its representation by querying the joint context formed by the other two modalities, thereby explicitly modeling conditional and bidirectional cross-modal interactions. 

Specifically, IMM consists of three parallel cross-attention components. In each component, one modality serves as the query, while the remaining two modalities are concatenated along the temporal dimension and used jointly as the key–value pair.
One representative component (the rightmost box in Fig. \ref{fig:archiFig}) is defined as follows:
\begin{equation}
\mathbf{Q}_{S} = \hat{h}_t^{s}, \quad  \mathbf{K}_{AP} = \mathbf{V}_{AP} =
[x_t^a|| h_t^{p}].
\end{equation}
Cross-attention is then applied to allow the proprioceptive representation to attend to the joint audio–geometric context:
\begin{equation}
\mathbf{Z}_{S\leftarrow AP}
= \mathrm{CrossAttn}(\mathbf{Q}_{S}, \mathbf{K}_{AP}, \mathbf{V}_{AP}).
\end{equation}
A residual connection followed by layer normalization is used to stabilize learning:
\begin{equation}
\widetilde{\mathbf{Z}}_{S\leftarrow AP}
= \mathrm{LayerNorm}\left(
\mathbf{Z}_{S\leftarrow AP} + \mathbf{Q}_{S}
\right).
\end{equation}

The same fusion structure is applied to the other two modality pairs, yielding $\widetilde{\mathbf{Z}}_{A\leftarrow PS}$ and $\widetilde{\mathbf{Z}}_{P\leftarrow AS}$, where each modality attends to the combined information from the other two.
Finally, the three cross-attended representations are concatenated and projected into a unified interaction-aware embedding:
\begin{equation}
\mathbf{z}_t
= \left[
\widetilde{\mathbf{Z}}_{A\leftarrow PS}
\;\Vert\;
\widetilde{\mathbf{Z}}_{P\leftarrow AS}
\;\Vert\;
\widetilde{\mathbf{Z}}_{S\leftarrow AP}
\right]
\end{equation}

This design enables explicit cross-modal reasoning by allowing each modality to attend to a joint cross-modal context formed by the other two. 
As a result, the learned representation $z_t$ captures interaction-level dependencies with an emphasis on audio features, leading to more informed and robust decision-making in multimodal manipulation tasks.
Finally, $z_t$ is used as a conditioning variable to generate actions in an end-to-end manner as introduced in Section \ref{pre}.

\begin{figure}[!t]
\centering

\includegraphics[width=0.9\linewidth]{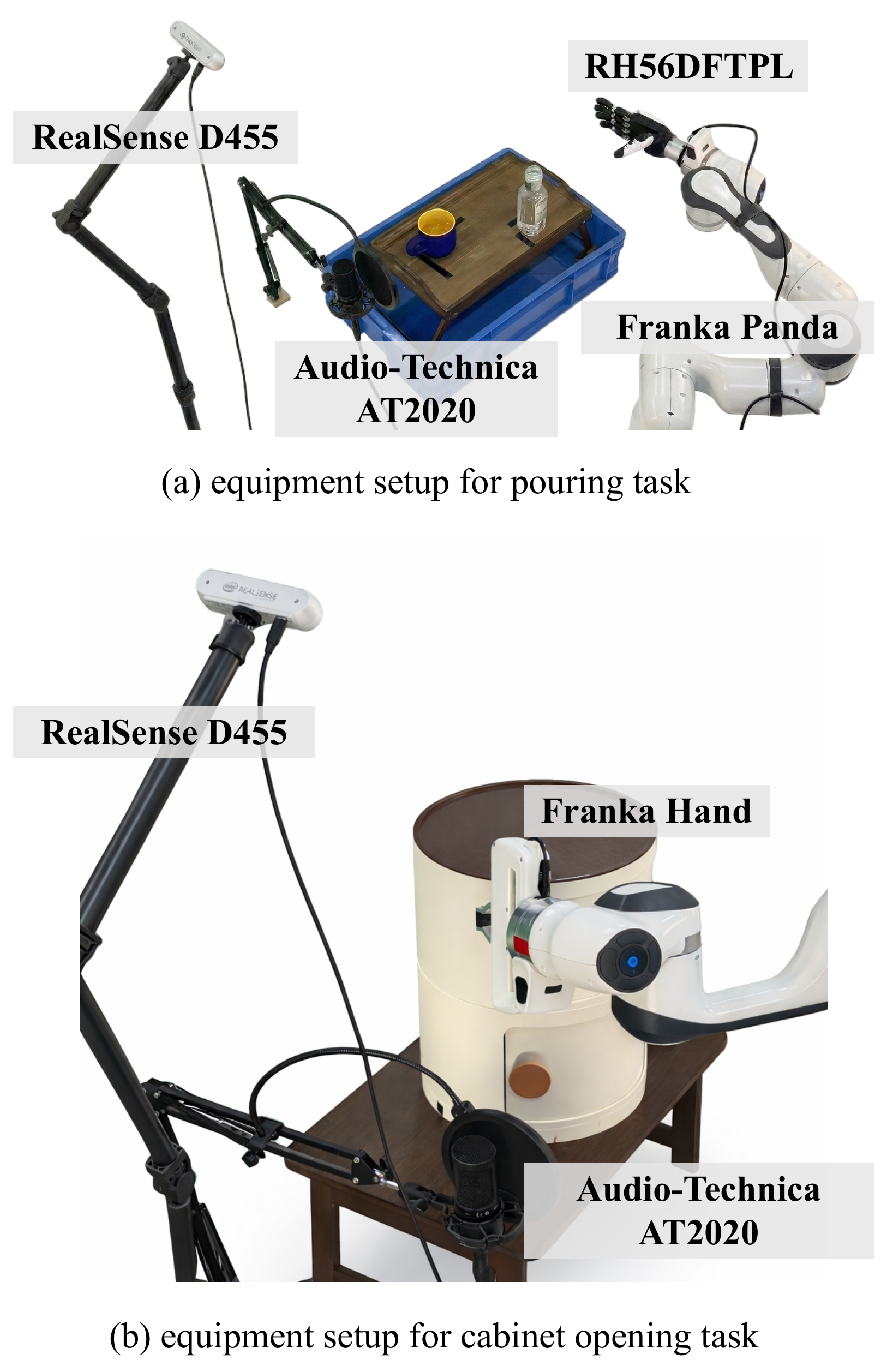}
\caption{\textbf{Overview of the experimental environment.} Both experiments involve visual, acoustic, and proprioceptive sensing.}
\label{fig:setup}
\end{figure}
\section{Experiments}
We evaluate the proposed method through a series of experiments. These experiments are designed to answer the following
key research questions:
\begin{itemize}
    \item Can audio feedback improve precision in acoustic-rich manipulation tasks? (Section~\ref{exp:comp})
    \item How does the proposed hierarchical fusion framework compare to existing multimodal methods? (Section~\ref{exp:comp})
    \item Can the proposed approach enhance the generalization ability of manipulation learning? (Section~\ref{exp:gen})
    \item Which component in the hierarchical fusion structure is more important? (Section~\ref{exp:ablation})
    \item Does the hierarchical fusion method preserve more task-relevant information? (Section~\ref{exp:mi})
\end{itemize}
\subsection{Setup}
\label{exp:setup}

We evaluate the proposed hierarchical multimodal fusion entirely in real-world settings using a Franka Emika Panda robot arm. 
As shown in Fig.~\ref{fig:setup}, the robot is equipped with an Inspire RH56DFTPL dexterous hand for pouring and a Franka gripper for the cabinet opening task. 
The system adopts a dual-computer architecture. A NUC is directly connected to the robot controller and handles low-level control, while a separate policy workstation acquires multimodal observations (audio, vision, and proprioception), executes the learned diffusion policy, and transmits predicted actions to the NUC for execution. 

\noindent {\bf Audio sensing} is collected using an Audio-Technica AT2020 microphone with a TASCAM US-1x2HR audio interface, positioned to capture interaction sounds within the workspace.

\noindent {\bf Visual observation} is obtained from an Intel RealSense D455 depth camera mounted above the workspace, providing a stable top-down view.

\noindent {\bf Robot proprioception} is accessed via the \texttt{panda\_py} interface. 
For the pouring task, proprioception consists of the end-effector Cartesian position and orientation, while for cabinet opening it additionally includes the gripper closure state of the Franka gripper.

\subsection{Baselines and Metrics}
\label{exp:baseline}
\subsubsection{Baselines}
The proposed method is compared with both a vision-based method and multimodal manipulation methods:

\noindent {\bf PointCloud-State ($PC \;\Vert\; State$).} This baseline directly concatenates point cloud features with robot proprioceptive features, without using audio information~\cite{Ze2024DP3}.

\noindent {\bf Audio-PointCloud-State ($Audio\;\Vert\;PC\;\Vert\;State$).} This naive multimodal baseline concatenates all three modality features straight, following the approach in~\cite{Ye-MORPHeus-2024}.

\noindent{\bf ManiWAV Fusion.} This baseline adopts an implicit multimodal fusion. Audio and visual features are first passed through a Transformer encoder, then concatenated with proprioception and processed by a multilayer perceptron to produce the final fused representation, as described in \cite{pmlr-v270-liu25c}.

\subsubsection{Evaluation Metrics}
\begin{figure}[!t]
\centering
\includegraphics[width=\linewidth]{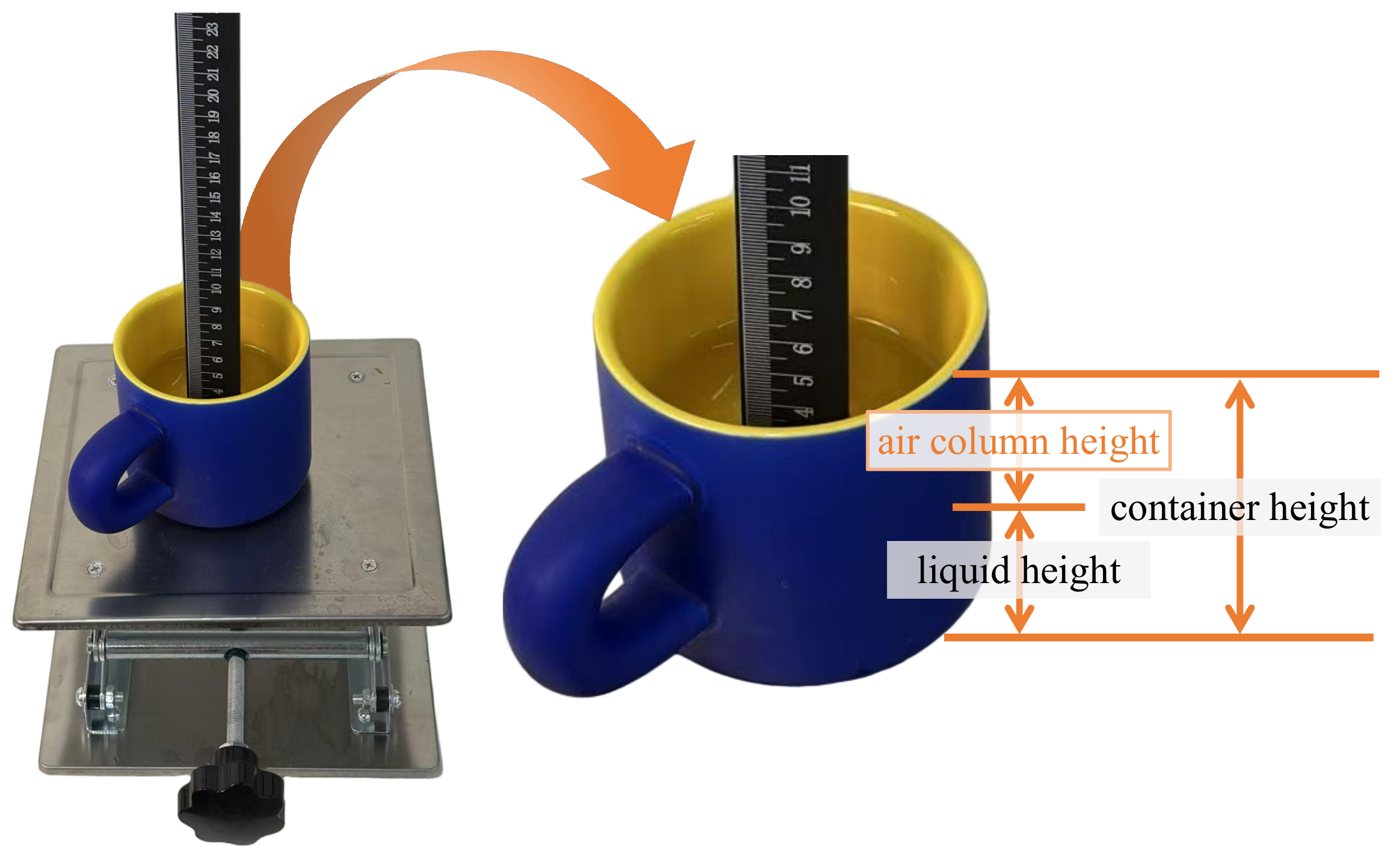}
\caption{{\bf Measurement of the liquid level.} We measure the height of the liquid after the robot arm has poured. The air column height is then computed as the container height minus the measured liquid height.} 
\label{fig:liquidMeasure}
\end{figure}

The policy performance is evaluated using the following precision-related metrics:

\noindent \textbf{Pouring Task.}
We evaluate manipulation precision using the height of the air column in the target container after pouring. 
At the beginning of each trial, the target container is initialized with an identical air column height. 
The robot then pours water from a source container into the target container. 
Performance is quantified by the final air column height in the target container. 
Since the air column height equals the container height minus the liquid height, this metric directly reflects the final liquid level achieved by the robot. 
To obtain this measurement, the target container is placed on a height-adjustable platform, as illustrated in Fig.~\ref{fig:liquidMeasure}. 
After pouring, the platform is gradually raised until the liquid surface contacts a vertically suspended ruler, allowing the liquid height to be measured accurately. 
A lower final air column height (i.e., a higher remaining liquid level) indicates better pouring performance. 

\begin{figure}[!t]
\centering
\includegraphics[width=\linewidth]{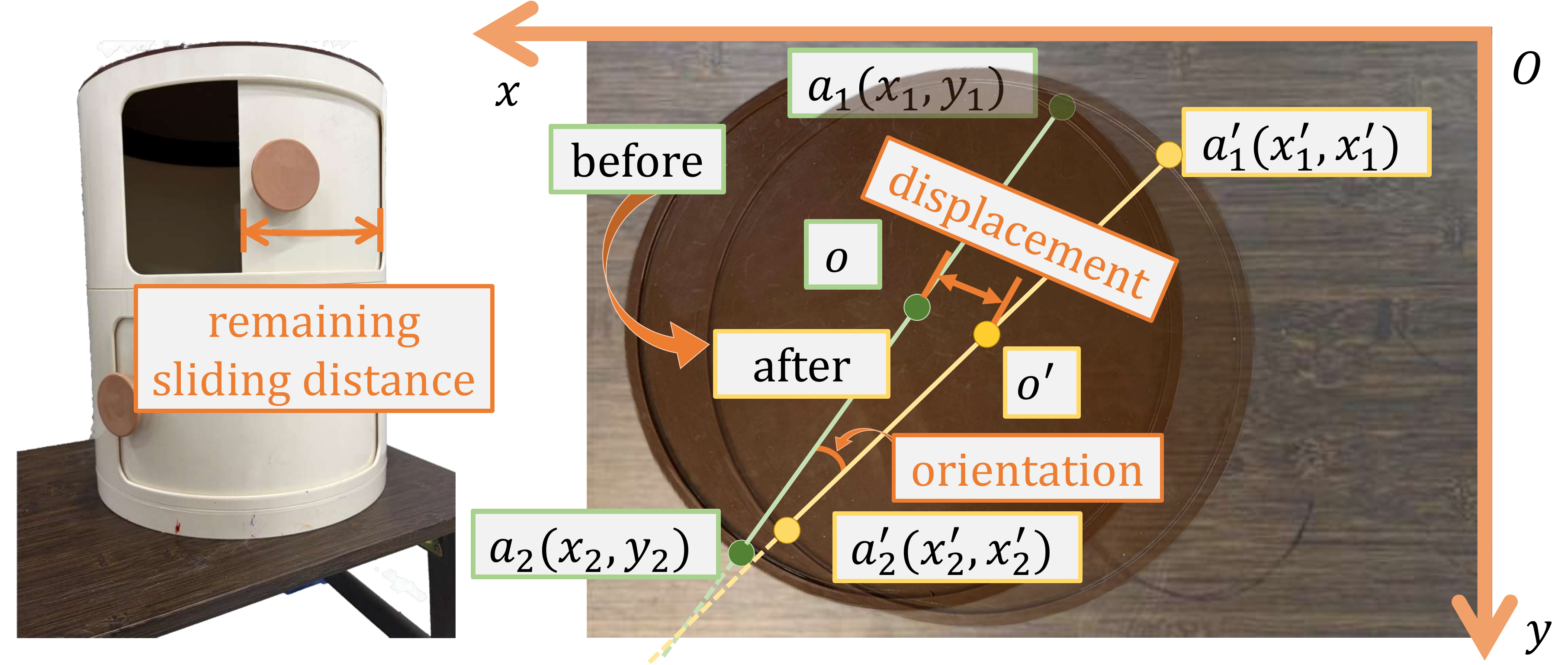}
\caption{{\bf Measurement in the cabinet opening task.} The left is the measurement of the remaining sliding distance of the cabinet door. The right is a top view of the cabinet on the desk. The coordinates of two reference points on the cabinet base diameter are recorded before (green) and after (orange) the gripper pulls the door, enabling computation of the cabinet’s translational displacement and rotational change.}
\label{fig:opening_measure}
\end{figure}

\noindent\textbf{Cabinet Opening Task.}
To evaluate the stability and accuracy of cabinet manipulation, we assess task performance using a composite score that accounts for both task completion quality and unintended cabinet motion.
The cabinet door is initially closed. The robot gripper first opens the door by pulling along the curved edge of the cabinet and subsequently closes it. 
While the door handle is visible at the beginning of the motion, its final position lies outside the camera view. 
Consequently, task completion is inferred from the acoustic cue generated by the final contact when the door reaches its closed position. 
To quantify cabinet motion induced during the interaction, we record the coordinates of two reference points on the circular base of the cabinet in the desk coordinate frame, both before and after the closing action. 
These measurements are used to compute the translational displacement and the rotational change of the cabinet, as illustrated in Fig.~\ref{fig:opening_measure}. 
Performance is evaluated using a weighted combination of three terms: the remaining sliding distance of the door $d_{\text{slide}}$, the translational displacement of the cabinet $d_{\text{disp}}$, and the rotational change of the cabinet $\theta_{\text{rot}}$. 
The overall score is computed as
\begin{equation}
\label{eq:cabinetmetric}
    \mathcal{S}
    =
    \alpha\, d_{\text{slide}}
    +
    \beta\, d_{\text{disp}}
    +
    \gamma\, \theta_{\text{rot}}, 
\end{equation}
where $\alpha$, $\beta$, and $\gamma$ denote the weighting coefficients. 
In our experiments, we set $\alpha = 0.3$, $\beta = 0.3$, and $\gamma = 0.4$, placing slightly higher emphasis on minimizing cabinet rotation. 
This composite metric captures both how completely the door is closed and how much the cabinet is disturbed during the manipulation process. 
Lower scores indicate better performance, corresponding to a fully closed door with minimal translational and rotational cabinet motion. 

\subsection{Comparison Results}
\label{exp:comp}




Table~\ref{tab:comparison} reports a quantitative comparison of different fusion methods on the pouring and cabinet opening tasks, where the results are reported as mean $\pm$ standard deviation over five trials.
Overall, the results show that incorporating audio as an additional modality can improve robotic manipulation performance, and our hierarchical fusion method significantly outperforms the baselines, including flat fusion and ManiWAV (Transformer-based) fusion. 

\input{tables/table1}

In the pouring task, the baseline method using only point cloud data and robot proprioception achieves an average final air-column height of 2.72\,cm. When audio is introduced through our hierarchical fusion approach, this value is reduced to 1.86\,cm, indicating that acoustic cues provide complementary information beyond visual geometry and robot state. A similar trend is observed in the cabinet opening task. The method without audio attains a score of 7.58, which is higher than the simple audio concatenation baseline (7.44) and substantially higher than both the ManiWAV fusion method (4.53) and our approach (4.50).

At the same time, the results indicate that adding audio alone does not guarantee improved performance. Methods based on flat concatenation or implicit Transformer-based fusion show only marginal gains and, in some cases, degrade performance. For example, in the pouring task, audio concatenation yields an air-column height of 3.08\,cm, comparable to ManiWAV-style fusion at 3.02\,cm, and both are worse than the baseline without audio. We attribute this behavior to the fact that acoustic signals in manipulation tasks are often intermittent and time-dependent. During large portions of the interaction, audio feedback may be weak or irrelevant, and without mechanisms to regulate cross-modal interactions, such signals can introduce noise rather than useful information.

In contrast, our approach explicitly constructs cross-modal interactions through a hierarchical fusion structure, allowing the policy to emphasize task-relevant acoustic cues while suppressing irrelevant ones. This leads to clear performance gains in audio-rich tasks such as pouring, consistent with the mutual information analysis in subsection~\ref{exp:mi}. For the cabinet opening task, where acoustic feedback is sparse and primarily occurs at the end of the interaction, the benefit of audio-aware fusion is less pronounced. Even so, our method matches the performance of ManiWAV fusion while exhibiting greater stability, suggesting that the value of advanced fusion strategies depends on how informative the audio modality is for the task.

Finally, our method consistently exhibits lower variance across trials, particularly in the cabinet opening task.
The reduced standard deviation indicates more stable and robust behavior, suggesting that hierarchical fusion helps mitigate the influence of noisy or weak modalities.

\subsection{Generalization Experiments}
\label{exp:gen}
\begin{figure}[!t]
\centering
\includegraphics[width=\linewidth]{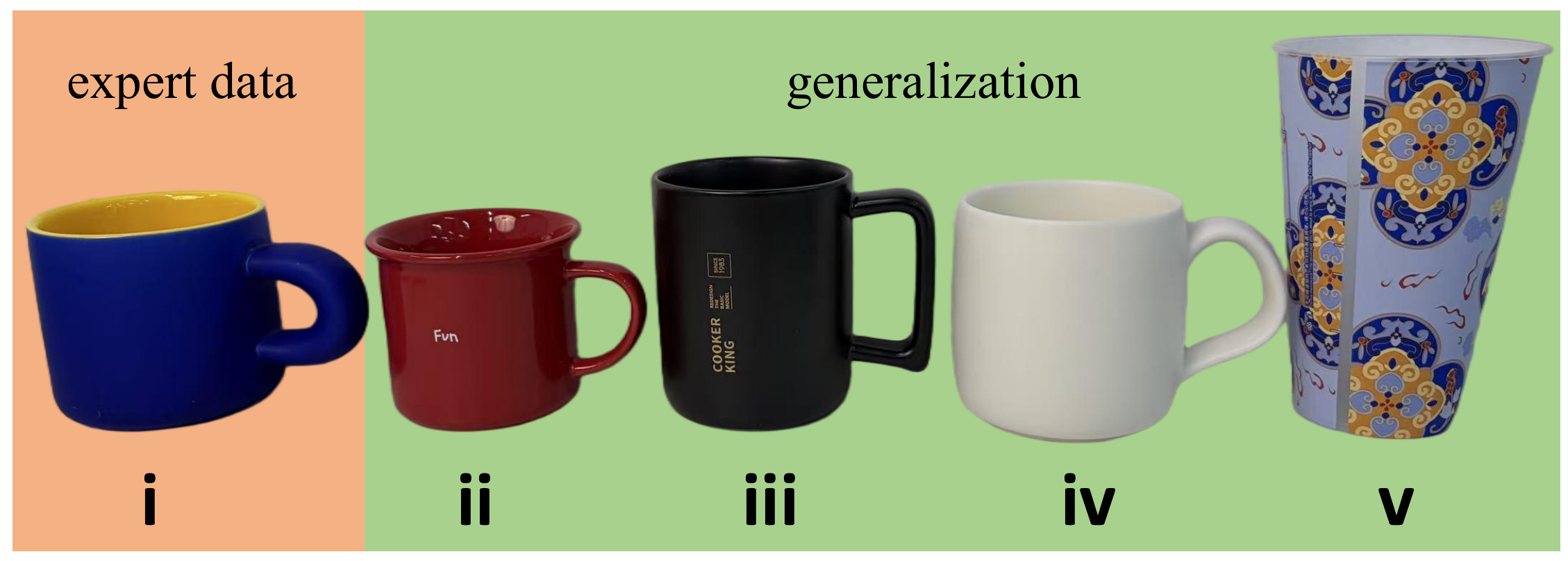}
\caption{\textbf{Target containers used in the pouring experiments.} Cup~(i) is used for expert data collection, training, and in-distribution evaluation in subsection~\ref{exp:comp}. Cups~(ii)–(v) are used exclusively for generalization experiments in subsection~\ref{exp:gen}.} 
\label{fig:cups}
\end{figure}
To evaluate generalization, we test the learned policy on four target containers with different geometry and appearance, as shown in Fig.~\ref{fig:cups}, while keeping the remaining task setup unchanged. All evaluations on Cups~(ii)–(v) are conducted in a zero-shot setting: the policy is trained exclusively on Cup~(i), with no additional data, fine-tuning, or adaptation for the unseen containers.
As shown in Table~\ref{tab:generation}, the performance of all methods degrades as the target container deviates further from the one used to collect expert demonstrations.
This trend is expected because the target container affects both the visual observations (e.g., shape and rim geometry) and the interaction acoustics (e.g., resonance and impact characteristics), thereby inducing a distribution shift in multimodal inputs.

Among the evaluated containers, the blue cup exhibits the largest domain gap due to its substantially different geometry (an inverted conical frustum rather than a straight cylinder), which leads to noticeably different audio signatures and altered fluid or impact dynamics.
Despite this increasing domain shift, our hierarchical fusion consistently achieves the best or comparable performance across all target containers.
These results suggest that explicitly structured multimodal fusion improves robustness to container-dependent variations and helps preserve task-relevant cues under out-of-distribution conditions.
Compared with implicit fusion (ManiWAV) and simple concatenation, our method retains higher performance under larger domain shifts, indicating that explicit hierarchical interactions better exploit complementary cues from audio, vision, and proprioception.

\begin{table}[t]
\centering
\caption{Generalization performance on novel target containers in the pouring task (mean $\pm$ std over 5 trials). Lower is better.}
\label{tab:generation}
\resizebox{\columnwidth}{!}{
\begin{tabular}{lcccc}
\toprule
\textbf{Method} 
& \textbf{red cup(ii)} 
& \textbf{black cup(iii)} 
& \textbf{white cup(iv)}
& \textbf{blue cup(v)}\\
\midrule
$PC\;\Vert\;State$
& $2.26 (\pm 0.19)$ 
& $2.6  (\pm 0.14)$ 
& $3.3  (\pm 0.09)$ 
& $3.42 (\pm 0.07)$ \\

$Audio\;\Vert\;PC\;\Vert\;State$
& $3.08 (\pm 0.23)$ 
& $3.12 (\pm 0.07)$ 
& $3.4  (\pm 0.13)$ 
& $3.68 (\pm 0.12)$ \\

ManiWAV Fusion 
& $2.3  (\pm 0.18) $ 
& $2.6  (\pm 0.13) $ 
& $3.32 (\pm 0.17) $ 
& $3.58 (\pm 0.26) $ \\

\textbf{Ours (Hierarchical)} 
& $\mathbf{1.78 (\pm 0.25)}$ 
& $\mathbf{2.5  (\pm 0.06)}$ 
& $\mathbf{2.8  (\pm 0.17)}$ 
& $\mathbf{3.14 (\pm 0.14)}$ \\
\bottomrule
\end{tabular}}
\end{table}

\subsection{Ablation Studies}
\label{exp:ablation}
Our hierarchical framework consists of two main components: (i) B-BFM, which conditions audio information onto point cloud and robot proprioceptive representations, and (ii) IMM, which models cross-modal relationships through attention over the conditioned features with an audio highway. To understand the contribution of each component, we conduct an ablation study on both manipulation tasks. The results are summarized in Table~\ref{tab:ablation}.

In the pouring task, using only the Binary-Branched Fusion module results in poor performance, indicating that conditioning audio onto visual and proprioceptive features alone is insufficient to guide accurate pouring. The model with only the Interaction Modeling module performs better than this baseline but still falls short of the full model, suggesting that interaction modeling without structured audio conditioning cannot fully capture the task dynamics.
A similar trend is observed in the cabinet opening task. While the Binary-Branched Aggregation Fusion–only variant slightly outperforms the Interaction Representation Learning–only variant, both exhibit significantly worse performance than the complete framework. This indicates that neither component alone can reliably handle the interaction complexity of the task.

The ablation results demonstrate that the two components are complementary. Effective performance is achieved only when audio is first used to condition visual and proprioceptive representations, and these conditioned features are then jointly processed by the interaction Modeling module. This combination enables the model to exploit audio cues while reasoning explicitly over all three modalities, leading to the best results across both tasks.

\begin{table}[htbp]
\centering
\caption{ablation studies for the hierarchical framework.}
\label{tab:ablation}
\begin{tabular}{lcc}
\toprule
\textbf{Method} & Pouring & Cabinet Opening \\
\midrule
B-BFM Only 
& $4.60 (\pm 0.00)$ 
& $6.21 (\pm 2.29)$ 
 \\ 	 
IMM Only 
& $3.38 (\pm 0.47)$
& $6.70 (\pm 1.46)$ \\  	 	 	 
\textbf{Ours}
& $\mathbf{1.86 (\pm 0.13)}$ 
& $\mathbf{4.50 (\pm 0.54)}$ \\  
\bottomrule  
\end{tabular}
\end{table}

\subsection{Mutual Information Analysis}
\label{exp:mi}
We employ mutual information (MI) to quantify how much task-relevant information is preserved in the fused multimodal representation.
Specifically, we measure the mutual information between the fused latent representation $\mathbf{z}_t$ and the task evaluation metric $y$, denoted as $I(\mathbf{z}_t; y)$, which characterizes how much uncertainty about $y$ is reduced by observing the multimodal representation.
\begin{equation}
I(\mathbf{z}_t; y)
= \mathbb{E}_{p(\mathbf{z}_t, y)}
\left[
\log \frac{p(\mathbf{z}_t, y)}{p(\mathbf{z}_t)p(y)}
\right]. 
\label{eq:mi_fused}
\end{equation}
In this context, a higher $I(\mathbf{z}_t; y)$ indicates that the representation fuses information that is predictive of the task outcome, rather than merely correlating with expert actions.
For the pouring task, the target variable $y$ is the final poured liquid amount, while for the cabinet-closing task, $y$ is the task score as defined in Equation~\ref{eq:cabinetmetric}.
\begin{table}[t]
\centering
\caption{$I(\mathbf{z}_t; y)$ analysis on pouring and opening cabinet tasks}
\label{tab:mi_pour}
\resizebox{\columnwidth}{!}{
\begin{tabular}{lcc}
\toprule
\textbf{Method} & $I(\mathbf{z}_t; y)_{pouring}$ & $I(\mathbf{z}_t; y)_{cabinet}$  \\
\midrule
$PC\;\Vert\;State$   & 0.039    & 0.047\\
$Audio\;\Vert\;PC\;\Vert\;State$ & 0.082 & 0.071 \\
ManiWAV Fusion  & 0.041   & 0.041\\
\textbf{Ours} & \textbf{0.088}   & \textbf{0.097}\\
\bottomrule
\end{tabular}}
\end{table}
Table~\ref{tab:mi_pour} summarizes the mutual information $I(\mathbf{z}_t; y)$ between the fused representation and the task outcome for both the pouring and cabinet-closing tasks.
Across both tasks, incorporating audio consistently increases the MI compared with the $PC\parallel State$ baseline, indicating that acoustic cues provide complementary task-relevant information beyond geometric and proprioceptive observations.

Compared with flat concatenation and ManiWAV fusion, our hierarchical method achieves the highest MI in both tasks.
In the pouring task, our fused representation yields an MI of 0.088, surpassing both flat concatenation (0.082) and ManiWAV (0.041), suggesting that the proposed hierarchical fusion more effectively preserves task-relevant information related to liquid dynamics.
A similar trend is observed in the cabinet-closing task, where our method attains an MI of 0.097, significantly higher than flat baselines and ManiWAV fusion.

\section{Conclusion and Future Work}
This paper examines the role of audio in multimodal robotic manipulation and introduces an explicit hierarchical fusion framework that integrates audio, visual, and robot proprioceptive information within an end-to-end diffusion policy. Real-world experimental results demonstrate that the proposed approach consistently outperforms flat concatenation and Transformer-based fusion methods. 
Nevertheless, the proposed framework has limitations. The benefits of hierarchical audio fusion are most evident in acoustically rich tasks, while performance gains are less pronounced when audio feedback is sparse or task-irrelevant. This suggests that the future development of effective multimodal fusion should account for the task relevance of each sensory modality.


\bibliographystyle{plainnat}
\bibliography{references}

\end{document}

%% file: tables/table1.tex
\begin{table}[htbp]
\centering
\caption{Performance comparison across different fusion methods.}
\label{tab:comparison}
\resizebox{\columnwidth}{!}{
\begin{tabular}{lcc}
\toprule
\textbf{Method} 
& \textbf{Pouring} 
& \textbf{Cabinet Opening} \\
\midrule
$PC \;\Vert\; State$
& $2.72 (\pm 0.07)$ 
& $7.58 (\pm 5.39)$ \\

$Audio\;\Vert\;PC\;\Vert\;State$
& $3.08 (\pm 0.13)$ 
& $7.44 (\pm 4.23)$ \\

ManiWAV Fusion 
& $3.02 (\pm 0.07)$ 
& $4.53 (\pm 1.37)$ \\

\textbf{Ours (Hierarchical)} 
& $\mathbf{1.86 (\pm 0.13)}$ 
& $\mathbf{4.50 (\pm 0.54)}$ \\
\bottomrule
\end{tabular}}
\end{table}